\documentclass[conference]{IEEEtran}
\IEEEoverridecommandlockouts

\usepackage{cite}
\usepackage[cmex10]{amsmath}
\usepackage{amsmath,amssymb,amsfonts}
\usepackage{algorithm}
\usepackage{algorithmic}
\usepackage{graphicx}
\usepackage{textcomp}
\usepackage{color}
\usepackage[table]{xcolor}

\usepackage{pifont}

\usepackage{enumitem}
\usepackage{path}
\usepackage{verbatim}
\usepackage{framed}
\usepackage{footnote}
\usepackage{cases}
\usepackage{subfigure}
\usepackage{theorem}
\usepackage{xurl}
\usepackage{outlines}
\usepackage{makecell}

\usepackage{acro}
\usepackage{enumitem}
\DeclareAcronym{ems}{
    short = EMS,
    long  = Electric Mobility Service,
}
\DeclareAcronym{ml}{
    short = ML,
    long  = Machine Learning,
}
\DeclareAcronym{dl}{
    short = DL,
    long  = Deep Learning,
}
\DeclareAcronym{mlr}{
    short = MLR,
    long  = Multiple Linear Regression,
}
\DeclareAcronym{lr}{
    short = LR,
    long  = Linear Regression,
}
\DeclareAcronym{dt}{
    short = DT,
    long  = Decision Tree,
}
\DeclareAcronym{rf}{
    short = RF,
    long  = Random Forest,
}
\DeclareAcronym{svm}{
    short = SVM,
    long  = Support Vector Machine,
}
\DeclareAcronym{svr}{
    short = SVR,
    long  = Support Vector Regression,
}
\DeclareAcronym{xgb}{
    short = XGB,
    long  = eXtreme Gradient Boosting,
}
\DeclareAcronym{lstm}{
    short = LSTM,
    long  = Long Short-Term Memory,
}
\DeclareAcronym{cnn}{
    short = CNN,
    long  = Convolutional Neural Networks,
}
\DeclareAcronym{ann}{
    short = ANN,
    long  = Artificial Neural Networks,
}
\DeclareAcronym{lgbm}{
    short = LGBM,
    long  = Light Gradient Boosting Machine,
}
\DeclareAcronym{knn}{
    short = kNN,
    long  = k-Nearest Neighbor,
}
\DeclareAcronym{ev}{
    short = EV,
    long  = Electric Vehicle,
}
\DeclareAcronym{hev}{
    short = HEV,
    long  = Hybrid Electric Vehicle,
}
\DeclareAcronym{phev}{
    short = PHEV,
    long =  Plug-in Hybrid Electric Vehicle,
}
\DeclareAcronym{ai}{
    short = AI,
    long  = Artificial Intelligence,
}
\DeclareAcronym{dnn}{
    short = DNN,
    long  = Deep Neural Networks,
}
\DeclareAcronym{soc}{
    short = SoC,
    long  = State of Charge,
}
\DeclareAcronym{rnn}{
    short = RNN,
    long  = Recurrent Neural Networks,
}
\DeclareAcronym{ddpg}{
    short = DDPG,
    long  = Deep Deterministic Policy Gradient,
}
\DeclareAcronym{pmp}{
    short = PMP,
    long  = Pontryagin's Minimum Principle,
}
\DeclareAcronym{rl}{
    short = RL,
    long  = Reinforcement Learning,
}

\acsetup{
  list/template = description
}
\setlist[description]{align=left, leftmargin=2cm, style=nextline}

\usepackage{tabularx}
\usepackage{booktabs}
\usepackage[colorlinks=true, allcolors=black]{hyperref}
\usepackage{threeparttable}

\def\BibTeX{{\rm B\kern-.05em{\sc i\kern-.025em b}\kern-.08em
    T\kern-.1667em\lower.7ex\hbox{E}\kern-.125emX}}
\begin{document}


\title{A Review on AI Algorithms for Energy Management in E-Mobility Services}

\author{Sen Yan, Maqsood Hussain Shah, Ji Li, Noel O'Connor and Mingming Liu 
\thanks{S. Yan ({\it sen.yan5@mail.dcu.ie}), N. O'Connor ({\it noel.oconnor@dcu.ie}) and M. Liu ({\it mingming.liu@dcu.ie}) are with the School of Electronic Engineering and SFI Insight Centre for Data Analytics at Dublin City University, Ireland. M. H. Shah ({\it maqsood@nuaa.edu.cn}) is with the Nanjing University of Aeronautics and Astronautics, China. J. Li ({\it j.li.1@bham.ac.uk}) is with the Department of Mechanical Engineering, University of Birmingham, UK. This work is supported by Science Foundation Ireland under Grant No. \textit{21/FFP-P/10266} and \textit{SFI/12/RC/2289\_P2}.}}


\maketitle

\begin{abstract}
E-mobility, or electric mobility, has emerged as a pivotal solution to address pressing environmental and sustainability concerns in the transportation sector. The depletion of fossil fuels, escalating greenhouse gas emissions, and the imperative to combat climate change underscore the significance of transitioning to electric vehicles (EVs). This paper seeks to explore the potential of artificial intelligence (AI) in addressing various challenges related to effective energy management in e-mobility systems (EMS). These challenges encompass critical factors such as range anxiety, charge rate optimization, and the longevity of energy storage in EVs. By analyzing existing literature, we delve into the role that AI can play in tackling these challenges and enabling efficient energy management in EMS. Our objectives are twofold: to provide an overview of the current state-of-the-art in this research domain and propose effective avenues for future investigations. Through this analysis, we aim to contribute to the advancement of sustainable and efficient e-mobility solutions, shaping a greener and more sustainable future for transportation.
\end{abstract}


\printacronyms[
    heading=section*, 
    name=List of Abbreviations \& Acronyms, 
]

\section{Introduction}
\label{sec: intro}
\ac{ems} refers to the use of electric-powered vehicles, including E-bikes, E-scooters, \ac{hev}, and \ac{phev}, for transportation needs. \ac{ems} has rapidly transformed the transportation landscape, offering sustainable alternatives to traditional combustion engine vehicles. These \ac{ev}s not only address environmental concerns but also contribute to the development of an interconnected transportation ecosystem, advancing intelligent transportation systems (ITS). By embracing \ac{ems}, we promote a future where interconnected vehicles, advanced data analytics, and smart infrastructure combine to create a safer, more efficient, and sustainable transportation network.
Energy management is crucial in \ac{ems} to ensure the efficient operation of electric vehicles and their charging infrastructure. It involves controlling and optimizing energy flow to meet specific requirements. Three key concerns in \ac{ems} energy management include ensuring a reliable range (often referred to as range anxiety), optimizing charging rates, and maximizing energy storage lifespan. Achieving this requires coordinating electrical energy resources like charging stations, renewable energy sources, and energy storage systems to facilitate electric vehicle charging.

Effective energy management is crucial for multiple reasons. One important aspect is ensuring the availability of charging infrastructure to meet the rising demand for electric vehicle charging \cite{Zhang2020}. As the number of \ac{ev}s continues to grow, the charging load on the power grid can become substantial. Therefore, meticulous management is essential to prevent overloading the grid and potential blackouts. Another key benefit of energy management is optimizing the utilization of energy resources, minimizing wastage, and maximizing the efficiency of the charging process. This not only helps reduce operational costs but also enhances the overall sustainability of \ac{ems} systems. Additionally, energy management facilitates grid integration and empowers \ac{ev}s to contribute to the grid by providing ancillary services or participating in vehicle-to-grid systems, thus strengthening the grid's stability and responsiveness.

\ac{ai} technologies offer a transformative solution to the limitations of traditional energy management techniques in \ac{ems} \cite{Yang2020_intro}. Conventional methods, which are primarily based on predetermined charging schedules and basic load balancing algorithms, struggle to meet the dynamic optimization requirements and growing complexity of modern \ac{ems} \cite{Nigro2021}. In contrast, \ac{ai} leverages advanced algorithms and real-time data analysis to optimize charging strategies intelligently. By adapting to changing conditions, utilizing predictive modeling, and employing multi-objective optimization, \ac{ai} enables more efficient and effective energy management in \ac{ems}, addressing the demand for optimal charging solutions.

The potential of \ac{ai} in transforming energy management for \ac{ems} lies in its computational techniques, including \ac{ml} and \ac{dl}. \ac{ai} algorithms and data-driven approaches enable intelligent systems to adapt to varying conditions, optimize charging operations, predict user behavior, and manage energy resources in real time. \ac{ai} facilitates dynamic load balancing, efficient energy allocation, and demand-response strategies, resulting in improved charging infrastructure utilization, reduced energy costs, and enhanced grid integration. This paper comprehensively reviews \ac{ai} technologies and techniques for energy management in \ac{ems}, covering energy consumption modeling, estimation, and prediction. It also discusses current challenges and proposes a research roadmap for future advancements. By assessing the state of \ac{ai}-based energy management, this paper contributes to the development of effective and sustainable \ac{ems} solutions.

The paper is organized as follows.
\autoref{sec: method} presents the methodology used in our paper, proposes the research questions we plan to investigate and summarizes and compares other existing surveys. \autoref{sec: conventional} provides an overview of conventional energy management systems, discussing their advantages and limitations.  \autoref{sec: AI} focuses on \ac{ai} approaches for energy management, delving into the current state of affairs in this domain. \autoref{sec: discussion} provides some discussion and introduces challenges to \ac{ai}-based energy management methods. Finally, \autoref{sec: conclusion} offers a brief conclusion summarizing the entire paper and presents future research directions.

\section{Review Methodology}
\label{sec: method}
The literature survey process was executed meticulously in five distinct phases, namely planning, literature research, reporting and interpretation of findings, and the synthesis of challenges and potential research directions for the future. This section provides a comprehensive account of the pivotal research questions to be explored and expounds upon the systematic methodology employed in conducting the literature search.

    \subsection{Research Questions}

    This paper aims at answering the following questions in relation to the application of \ac{ai} methods in \ac{ems}:
    
    \begin{enumerate}
        \item What are the existing \ac{ai} technologies and techniques used for energy management in \ac{ems}?
        \item How are \ac{ai}-based approaches employed in energy consumption modeling, estimation, and prediction in \ac{ems}?
        \item What are the current challenges and limitations of \ac{ai} methods in energy management for \ac{ems}?
        \item What is the future research roadmap for advancements in \ac{ai}-based energy management for \ac{ems}?
        \item How does the use and focus of \ac{ai} approaches vary among different \ac{ems}?
    \end{enumerate}
    
    \subsection{Literature Retrieval}

    We conducted a systematic search of peer-reviewed research publications to collect studies that employed \ac{ai} approaches to address issues related to energy management in \ac{ems}. Our screening process involved a thorough review of the literature to identify papers that addressed the structural challenges of \ac{ems} energy management and utilized \ac{ai} methods. We utilized reputable online databases, including Google Scholar, ACM Digital Library, Springer, MDPI, IEEE, and Science Direct, which index a wide range of computer science and technology research, to ensure comprehensive coverage of relevant studies.

    The literature search process was conducted using a set of specific keywords, including ``energy management", ``electric mobility service", ``machine learning", ``EV", ``e-bike", ``e-scooter" and "energy consumption prediction". Only research papers written in English were included in the search. As a result of this comprehensive search, a total of approximately 30 papers were retrieved for review. Among these papers, 1 of them specifically focused on E-scooters \cite{na2022}, 1 paper focused on E-bikes \cite{Burani2022}, and the remaining papers centered on EVs.

    The results show that few existing survey papers in the literature focused on the applications of \ac{ai} methods for energy management in E-micromobility systems, such as E-bikes and E-scooters. All selected papers for review are relevant in our context, which highlights the applicability of \ac{ai}-based approaches in dealing with energy consumption prediction problems in different aspects.
    
    \subsection{Existing Survey}

    In this section, we provide a comprehensive summary and comparison of existing surveys pertaining to energy management, e.g., the estimation of battery \ac{soc}, in \ac{ems}. It is evident that the field commonly accepts the use of three main categories of estimation approaches: electrochemical models, equivalent circuit models, and data-driven models. However, in recent years (2019 to 2022), there has been a notable emphasis on ``data-driven methods" (such as \ac{ai} approaches) and ``connected environments" in the future direction section of these surveys. This highlights the growing importance and attention given by researchers to these areas in energy management systems.

    Given this context, our work primarily aims to summarize the various modeling, estimation, and prediction approaches utilized in this domain. The objective is to provide readers with a concise understanding of the available models or algorithms and offer suggestions for their appropriate selection based on different cases and scenarios. By offering this overview, we aim to assist researchers and practitioners in making informed decisions regarding the most suitable approaches for their specific energy management requirements within \ac{ems}.
    
    \begin{table*}[ht]
        \caption{Existing surveys focusing on energy management in \ac{ems}.}
        \label{table: survey}
        \begin{tabularx}{\linewidth}{@{\extracolsep{\fill}}c l l l l c}
            \toprule
            \textbf{Object} & \textbf{Year} & \textbf{Method(s) reviewed} & \textbf{Discussion field(s)} & \textbf{Future direction(s)} & \textbf{Ref}\\
            \midrule
            \makecell{\ac{hev} \& \\\ac{phev}}  & 2019 & \ac{ai} methods & \makecell[l]{Energy management problems \\\ac{rl} layouts \\ Applications of \ac{rl} approaches} & \makecell[l]{Novel \ac{rl} algorithms \\Energy management in intelligent transportation systems \\ Multi-objective optimization \\Cooperative learning in a connected environment} & \cite{Hu2019}\\
            \midrule
            \makecell{\ac{hev} \& \\\ac{phev}}  & 2019 & \makecell[l]{Conventional methods \\\ac{ai} methods} & \makecell[l]{\ac{hev} \& \ac{phev} definitions \\Classification of strategies} & \makecell[l]{Connected \& automated systems \\ Driver-in-the loop systems \\Multi-objective optimization \\Learning-based methods \\Cooperative adaptive cruise controller-based systems \\Multi-scale systems \\ Cooperative learning in a connected environment} & \cite{Zhang2019}\\
            \midrule
            \ac{ev} & 2021 & \makecell[l]{Conventional methods \\\ac{ai} methods \\Hybrid methods} & \makecell[l]{Influential variables \\Modeling scale \\Modeling methodology} & \makecell[l]{Energy estimation for vehicles \\Application for vehicle-to-grid integration \\Multi-scale systems} & \cite{Chen2021}\\
            \midrule
            \ac{ev} & 2021 & \makecell[l]{Conventional methods \\\ac{ai} methods} & \makecell[l]{Definition of \ac{soc} \& state of health \\Novel estimation methods} & \makecell[l]{Estimation errors \\Gaps between lab and practice \\Joint estimation \\Different applications \\Data-driven method} & \cite{Wang2021}\\
            \midrule
            \ac{ev} & 2021 & \makecell[l]{Conventional methods \\\ac{ai} methods} & \makecell[l]{Battery modeling \\\ac{soc} estimation \\Battery charging} & \makecell[l]{Joint estimation \\Scalable state estimation \\Smart learning and optimization \\Safety Management} & \cite{Adaikkappan2021}\\
            \midrule
            \ac{ev} & 2022 & \makecell[l]{Conventional methods \\\ac{ai} methods} & \makecell[l]{Novel battery technologies \\Battery management technologies} & \makecell[l]{Batteries \\Technologies regarding batteries \\Technologies replacing batteries} & \cite{Liu2022}\\
            \midrule
            \textbf{\ac{ems}} & \textbf{2023} & \makecell[l]{\textbf{Conventional methods} \\\textbf{\ac{ai} methods}} & \makecell[l]{\textbf{Conventional estimation methods} \\\textbf{Novel \ac{ai}-based estimation methods} \\\textbf{Existing relevant surveys}} & \makecell[l]{\textbf{Data availability} \\\textbf{Model complexity} \\\textbf{Real-time prediction capabilities} \\\textbf{Integration with renewable energy sources} \\\textbf{Management of uncertainty and risk factors}} & \textbf{*}\\
            \bottomrule
        \end{tabularx}
        \begin{tablenotes}\footnotesize
            \item The last item marked in \textbf{bold} represents our work conducted in this paper.
        \end{tablenotes}
    \end{table*}

\section{Conventional Approaches}
\label{sec: conventional}
\begin{figure*}[ht]
    \vspace{-0.1in}
    \centering
    \includegraphics[width=\linewidth]{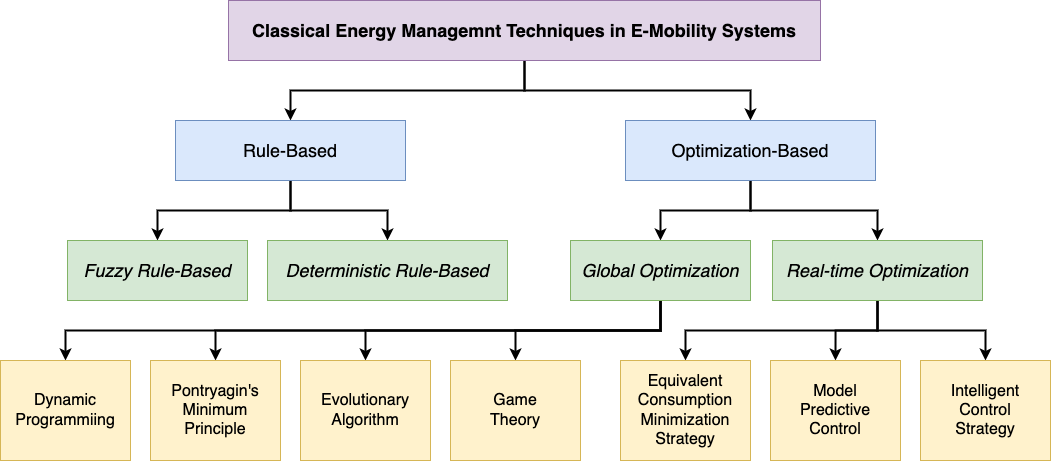}
    \caption{Hierarchical categorization of classical \ac{ems}.}
    \label{fig: classical ems}
    \vspace{-0.1in}
\end{figure*}

Based on the literature, conventional energy management methods for \ac{ev}s (\ac{hev}sor \ac{phev}s) can be classified into two main categories: rule-based and optimization-based. A brief summary of the advantages and limitations of conventional methods is provided in \autoref{table: conv}, and \autoref{fig: classical ems} shows the hierarchical categorization of classical energy management systems. 

Rule-based methods have been widely employed in early \ac{hev}s due to their simplicity and feasibility \cite{Anbaran2014}. These methods focus on coordinating the operation of the internal combustion engine to improve fuel economy and emission performance by transferring the working points of the engine from low to high-efficiency zones \cite{Anbaran2014}. Deterministic rule-based methods utilize heuristics, intuition, and mathematical models to develop control rules based on prior knowledge of the driving cycle \cite{Salmasi2007}. Fuzzy rule-based \ac{ems}, on the other hand, incorporates fuzzy logic control to enhance adaptability and robustness \cite{Wang2018}.

Optimization-based methods are categorized as global and real-time optimization methods. Various global optimization methods have been employed, including dynamic programming, \ac{pmp}, Evolutionary Algorithms, and Game Theory. Dynamic programming breaks down the decision process into discrete steps and has been used to solve the optimization problem of multi-step decision processes \cite{ChanChiaoLin2003}. \ac{pmp} finds optimal control signals for time-varying non-linear systems subject to constraints \cite{Xu2013}. Evolutionary Algorithms encompass swarm-based algorithms such as Particle Swarm Optimization and Genetic Algorithms \cite{Yang2020}. Game Theory treats the energy management problem as a game among decision-makers \cite{Dextreit2008}.

Real-time optimization methods aim at minimizing energy consumption dynamically and include methods such as Equivalent Consumption Minimization Strategy and Model Predictive Control. ECMS converts electric energy into equivalent fuel consumption, allowing for compromise optimization of the vehicle's dynamic performance, fuel economy, and emission performance \cite{kugor2014}. Model Predictive Control utilizes a prediction horizon and rolling optimization to determine optimal control actions in real-time \cite{Yang2020_con}.

As the complexity of energy management systems continue to rise, conventional approaches are being surpassed by more advanced AI methods, offering enhanced energy management capabilities.
\begin{table*}
    \caption{Summary of conventional methods in \ac{ems} energy management.}
    \label{table: conv}
    \begin{tabularx}{\linewidth}{@{\extracolsep{\fill}}l X X}
        \toprule
        \textbf{Method} & \textbf{Advantages} & \textbf{Limitations}\\
        \midrule
        Rule-based & \makecell[l]{Ease of implementation and feasibility \\Low computational burden} & \makecell[l]{Limited optimization and adaptability \\Inability to adapt to different conditions and dynamic changes}\\
        \midrule
        Fuzzy rule-based & \makecell[l]{Robustness and adaptability} & \makecell[l]{Difficulty in ensuring optimal control performance} \\
        \midrule
        Optimization-based & \makecell[l]{Global optimization capability \\Potential for multi-objective optimization \\Feasibility of real-time operation \\Applicability to different driving cycles \\Enhanced control performance and optimization effectiveness \\Adaptability to changing conditions and driver behaviors} & \makecell[l]{Dependence on prior knowledge of driving cycle conditions \\Heavy computational burden and longer computational time \\Inability to handle real-time requirements \\Uncertainty in driving conditions} \\
        \bottomrule
    \end{tabularx}
\end{table*}

\section{Artificial Intelligent Approaches}
\label{sec: AI}
In the realm of energy management for \ac{ems}, AI-based approaches emerge as markedly superior to conventional methods \cite{Yang2020_intro}. This distinction arises from the inherent attributes of \ac{ai} algorithms, which encompass dynamic adaptability, data-driven precision, and the capacity for continuous learning. Consequently, these algorithms possess the capability to process substantial real-time data and promptly adapt to changing circumstances, leading to effective energy consumption optimization \cite{Hu2023}. In the subsequent discussion, we review AI strategies deployed within \ac{ems} energy management, examining them from two vantage points: traditional \ac{ml} methods and \ac{dl} methods.

    \subsection{Traditional Machine Learning Methods}

    \ac{ml} methods leverage the inherent patterns present within the data to facilitate the learning and adaptation of the system, thereby enabling accurate predictions for previously unseen data. Various \ac{ml} approaches, including \ac{lr} \cite{Mediouni2022}, \ac{mlr} \cite{Ullah2021, Lpez2020, Abdelaty2021}, \ac{svm} or \ac{svr} \cite{Abdelaty2021, Ullah2021_esg, Ragone2021}, \ac{dt} \cite{Ullah2021_esg, Abdelaty2021}, \ac{rf} \cite{Ullah2021_esg, na2022, Li2021}, \ac{xgb} \cite{Ullah2021, Zhang2020}, \ac{lgbm} \cite{Ullah2021}, \ac{knn} \cite{Ullah2021_esg, na2022, Li2021}, and \ac{ann} \cite{Ullah2021_esg, Ullah2021, Abdelaty2021, Ragone2021}, have been widely employed to address the challenges associated with energy consumption modeling or prediction for \ac{ems}. A brief summary of the advantages and limitations of traditional \ac{ml} methods is provided in \autoref{table: ml}.

    \begin{table*}
        \caption{Summary of traditional \ac{ml} methods in \ac{ems} energy management.}
        \label{table: ml}
        \begin{tabularx}{\linewidth}{@{\extracolsep{\fill}}l X X}
            \toprule
            \textbf{Method} & \textbf{Advantages} & \textbf{Limitations}\\
            \midrule
            \ac{lr} \& \ac{mlr} & \makecell[l]{Ease of understanding and interoperability \\High computational efficiency \\Capability to handle multiple features \\Excellent performance on small datasets} & \makecell[l]{Limitation to linear relationships \\Assumption of data normality \\Sensitivity to outliers \\Issue of multicollinearity}\\
            \midrule
            \ac{svm} \& \ac{svr} & \makecell[l]{Ability to handle linear/non-linear classification problems \\Effectiveness even in situations with high feature dimensions} & \makecell[l]{Complexity of computations for large-scale data \\Need for parameter tuning \\Sensitivity to missing data} \\
            \midrule
            \ac{dt} \& \ac{rf} & \makecell[l]{Ease of understanding and interpretation \\Ability to handle numerical and categorical data \\Minimal need for data preprocessing} & \makecell[l]{Propensity for \ac{dt} to overfit \\Requirement for substantial computational resources for \ac{rf}} \\
            \midrule
            \ac{xgb} & \makecell[l]{High accuracy \\Prevention of computational resource waste \\Built-in handling of missing values \\Presence of regularization parameters to prevent overfitting} & \makecell[l]{Potential for long training times \\Sensitivity to parameter selection \\Potential need for extensive time in parameter tuning} \\
            \midrule
            \ac{lgbm} & \makecell[l]{Fast training speed \\Low memory usage \\High accuracy \\Capability to handle large-scale data} & \makecell[l]{Sensitivity to parameter selection \\Potential need for extensive time in parameter tuning} \\
            \midrule
            \ac{knn} & \makecell[l]{Simplicity and ease of understanding \\Insensitivity to outliers \\No data input assumptions} & \makecell[l]{Large computational requirement \\Need for substantial memory \\Poor performance on imbalanced sample issues} \\
            \midrule
            \ac{ann} & \makecell[l]{Capability to handle complex non-linear relationships \\Strong ability to process large-scale and high-dimensional data} & \makecell[l]{Need for large amounts of data for training \\Long training times \\Poor interpretability of the model} \\
            \bottomrule
        \end{tabularx}
    \end{table*}

    These \ac{ml} algorithms or models were individually applied and compared in various case studies. For example, \ac{mlr}, \ac{dt}, \ac{svm} and other neural network-based models were implemented on the data collected from electric buses in \cite{Abdelaty2021}. Furthermore, in \cite{Ullah2021}, the authors employed \ac{mlr}, \ac{ann}, \ac{xgb} and \ac{lgbm} to predict \ac{ev} energy consumption using a dataset collected in Japan. The results demonstrate the superiority of \ac{xgb} and \ac{lgbm} compared to other selected algorithms based on lower mean absolute error.
    
    On the other hand, combining \ac{ml} models may lead to improved performance. For instance, based on the combination of \ac{dt}, \ac{rf} and \ac{knn}, the authors designed a new method named Ensemble Stacked Generalization \cite{Ullah2021_esg} to predict the energy consumption of \ac{ev}s, and evaluated its performance on the same dataset collected in Japan. The results showed that, despite longer running times, the proposed method outperformed the baselines (i.e., \ac{dt}, \ac{rf} and \ac{knn}). Therefore, the authors concluded that adopting stacking techniques can enhance the accuracy of predictive models for \ac{ev} energy consumption.

    \subsection{Deep Learning Methods}

    \ac{dl} methods, a subset of \ac{ml}, possess the capability to extract intricate patterns from transportation data. In contrast to conventional \ac{ml} models, such as \ac{rf} and \ac{svm}, \ac{dl} models leverage neural network architectures comprising multiple hidden layers to capture intricate relationships within traffic big data. These \ac{dl} models excel at learning high-level representations of data, surpassing the limitations of human-designed features \cite{Gadri2021}.

    To illustrate the significance of \ac{dl} techniques, one can consider a representative scenario involving dynamic range optimization in electric fleet management. Traditional methods used to estimate the remaining operational range in \ac{ev} fleet management often rely on simplistic rules that struggle to accommodate real-world factors like traffic patterns and atmospheric conditions \cite{TROVAO2013304-R1}. In contrast, \ac{dl} methods offer a more advanced solution, leveraging their capacity to assimilate diverse features such as GPS data, weather conditions, and driver behaviors. This synthesis of data enables real-time adaptation of \ac{ev} range predictions  \cite{Zheng2022-R1}. Moreover, \ac{dl} can not only predict range more accurately and optimize it by suggesting efficient routes and driving modes but also personalize estimates for each driver to continuously improve the prediction performance based on collected data.
    
    To provide further specific examples of \ac{dl} models from the literature, researchers in \cite{How2020}, for instance, who employed various depths of \ac{dnn} models to estimate the \ac{soc} of \ac{ev} batteries. These models utilized open-circuit voltage measurements at different ambient temperatures as input variables. Moreover, \ac{rnn} models, including the \ac{lstm} variant, have found extensive utility in \ac{ems} due to their aptitude for capturing temporal dependencies in data. As presented in \cite{Hong2019}, \ac{lstm} was applied to predict multiple targets, e.g., voltage, temperature, and \ac{soc}, based on time series data, showcasing precise online prediction and robustness.
    
    Furthermore, \ac{cnn} models have found application in similar research areas by converting time series data into image representations. In particular, \cite{Modi2020} used the Gramian Angular Field approach to convert time series data into images, which were then fed into a CNN model to estimate \ac{ev} energy consumption. The CNN model's performance was evaluated against other baseline models such as \ac{ann} and \ac{mlr}.
    
    In the context of \ac{rl}, \cite{Hu2019} provided a comprehensive overview of various \ac{rl} methods. Two notable methods include \ac{ddpg} and a novel approach combining deep \ac{rl} with the \ac{pmp} method. In \cite{Qu2020}, authors introduced a \ac{ddpg}-based following model designed for connected and autonomous \ac{ev}s, aimed at mitigating traffic fluctuations caused by human drivers, known as stop-and-go traffic waves, while optimizing electrical energy consumption. Moreover, \cite{Hu2023} developed an electric management system that integrates deep \ac{rl} and the \ac{pmp} algorithm, demonstrating substantial performance improvements compared to traditional \ac{pmp}-based electric management systems.
    
    A brief summary of the advantages and limitations of \ac{dl} methods is provided in \autoref{table: dl} below.

    \begin{table*}
        \caption{Summary of \ac{dl} methods in \ac{ems} energy management.}
        \label{table: dl}
        \begin{tabularx}{\linewidth}{@{\extracolsep{\fill}}l X X}
            \toprule
            \textbf{Method} & \textbf{Advantages} & \textbf{Limitations}\\
            \midrule
            \ac{dnn} & \makecell[l]{Capability to handle complex non-linear relationships \\Strength in processing high-dimensional unstructured data \\Feature extraction and learning through hidden layers} & \makecell[l]{Requirement for large amounts of data for training \\Lower interpretability compared to some simpler \ac{ml} models \\High model complexity, requiring long training time}\\
            \midrule
            \ac{lstm} & \makecell[l]{Capability to handle long sequence data \\Resolution of the gradient issue in RNN} & \makecell[l]{Requirement for large amounts of data for training \\High model complexity, requiring long training time} \\
            \midrule
            \ac{cnn} & \makecell[l]{Suitability for handling high-dimensional input data \\Ability to automatically detect important features} & \makecell[l]{Requirement for large amounts of data for training \\Risk of overfitting \\ Requirement for approaches to convert time series to images} \\
            \midrule
            \ac{ddpg} & \makecell[l]{Ability to handle high-dimensional and continuous action spaces \\Capability to directly learn a policy \\Capability to stabilize the learning process} & \makecell[l]{Requirement for large amounts of data and training time \\Sensitivity to noise and outliers} \\
            \bottomrule
        \end{tabularx}
    \end{table*}

\section{Discussion \& Challenges}
\label{sec: discussion}
The future trajectory of energy consumption modeling, estimation, and prediction in the context of e-mobility unfolds a multitude of challenges that necessitate meticulous attention and innovative solutions to further propel this domain. These challenges encompass diverse dimensions, encapsulating data availability, model complexity, real-time prediction capabilities, integration with renewable energy sources, and the management of uncertainty and risk factors.
    \subsection{Data Availability}
    The availability of high-quality, real-world data plays a pivotal role in e-mobility energy consumption modeling. On the one hand, achieving the convergence of either the \ac{ml} model or the control policy is a protracted undertaking necessitating the acquisition of data from numerous iterative simulations \cite{Lee2021, HossainLipu2020}. On the other hand, many researchers have predominantly employed small-scaled datasets derived from conventional standardized driving cycles or constrained real-world driving scenarios to train predictive models. Such an approach potentially compromises the precision of these models when applied to authentic real-world driving contexts \cite{Tang2021, HossainLipu2020}. Therefore, it is imperative to acquire diverse and comprehensive datasets that encompass a wide range of driving conditions, vehicle types (corresponding to the extreme imbalance in E-bike, E-scooter and \ac{ev} systems), and user behaviors to ensure the accuracy and reliability of the models and predictions. Thus, efforts should be made to collect large-scale datasets with high granularity, including vehicle parameters (e.g., battery capacity and efficiency), driving patterns (e.g., speed profiles and acceleration patterns), and environmental factors (e.g., temperature and road conditions). Collaboration among researchers, industry partners, and policymakers can help overcome data accessibility and privacy concerns, allowing for the development of robust models.

    \subsection{Model Complexity\& Interpretability}

    With the increasing complexity of e-mobility systems, it is imperative that the models employed for energy consumption estimation and prediction possess the necessary capability to capture the dynamic nature and intricate interactions within the system. These models must be designed to be scalable, capable of handling large-scale deployments of \ac{ev}s, and adaptable to accommodate diverse vehicle types, environment factors, and user preferences.

    Advanced \ac{ai} methodologies, including \ac{dl} and \ac{rl}, offer avenues for the creation of intricate models adept at capturing nuanced interdependencies and nonlinear dynamics inherent to the system. Nevertheless, it is important to acknowledge that the elevated intricacy of these models will likely result in computational requisites that surpass the capabilities of the electronic control unit embedded within an operational vehicle powertrain, particularly when utilized as an online controller \cite{Lee2021, Sun2020, HossainLipu2020}.
    
    However, it should be noted that the interpretability of intricate models is relatively lower compared to physics-based methods. This is because data-driven methods rely on black-box models where detailed information is not known \cite{HossainLipu2020, Deng2020}. Thus, exploring hybrid models merging physics-based and data-driven techniques becomes relevant. These hybrids integrate fundamental \ac{ev} principles with data-driven methods, capturing real-world intricacies. Balancing accuracy and efficiency, they offer the potential for valuable insights.

    \subsection{Real-time Prediction}

    Real-time prediction of energy consumption is crucial for optimizing charging and discharging strategies, managing grid integration, and providing accurate range estimation to \ac{ev} users. However, achieving real-time predictions while considering dynamic factors such as traffic conditions, weather, and user behavior poses a significant challenge.

    Real-time prediction models can leverage techniques such as online learning, adaptive control, and model-based reinforcement learning. These approaches enable continuous learning from new data and allow for dynamic adaptation to changing conditions. Integration with real-time data sources, such as traffic information, weather forecasts, and vehicle-to-grid communication, can further enhance the accuracy of real-time predictions.

    \subsection{Integration with Renewable Energy Sources}

    Integrating e-mobility systems with renewable energy sources adds complexity to energy modeling and prediction. Renewable energy's intermittent nature and the need to balance supply and demand require precise predictions and optimization. Models must factor in the availability and variability of sources like solar and wind power, considering energy consumption patterns. We can employ techniques like probabilistic forecasting, optimization algorithms, and energy management systems to optimize renewable energy use, minimize grid strain, and reduce carbon emissions.

    \subsection{Uncertainty \& Risk Management}
    Uncertainties tied to factors like user behavior, charging infrastructure, and battery wear and tear present hurdles in accurately estimating and predicting energy consumption in e-mobility systems \cite{HossainLipu2020}.

    To tackle these uncertainties and their associated risks, we can employ probabilistic models, uncertainty quantification techniques, and risk analysis frameworks. Among these approaches, AI, particularly reinforcement learning (RL), stands out as a suitable solution. RL algorithms enable adaptive decision-making by utilizing real-time feedback, empowering e-mobility systems to optimize their energy management. For instance, RL can model user behavior to find optimal charging schedules based on preferences and past patterns, address battery wear by optimizing charging and discharging profiles, and optimize the use of charging infrastructure by allocating resources intelligently. By incorporating RL models, e-mobility systems can effectively handle uncertainties, boost operational efficiency, and ensure reliable performance.

    In summary, future challenges in e-mobility energy management include data availability and quality, model complexity and scalability, real-time prediction, renewable energy integration, and uncertainty management. Addressing these challenges demands interdisciplinary collaboration, advanced machine learning techniques, solid data infrastructure, and policy support to foster the development of dependable and efficient e-mobility systems.       

\section{Conclusion \& Future Directions}
\label{sec: conclusion}
In this survey paper, we meticulously examined the methods for modeling, estimating, and predicting energy consumption in electric mobility. We categorized these approaches into two main groups: conventional methods and AI-based algorithms. We synthesized insights from relevant surveys in this domain. Our analysis reveals significant progress in understanding energy consumption dynamics. Conventional methods provide foundational insights, while traditional machine learning algorithms excel in capturing patterns to make accurate predictions. Deep learning algorithms, on the other hand, excel in addressing intricate, non-linear dynamics. However, we also identify several challenges in this research area, including but not limited to the acquisition of diverse and high-quality datasets, addressing model complexity, achieving real-time predictions, integrating renewable energy sources, and effectively managing uncertainty. 


For future work, we recommend further exploring data-driven methods and real-time data integration to boost accuracy and performance. Additionally, the limited research in micro-mobility areas like E-bikes and E-scooters highlights the need for a thorough investigation in this domain.


\section*{Acknowledgment}

This work has emanated from research supported in part by Science Foundation Ireland under Grant Number \textit{21/FFP-P/10266} and \textit{SFI/12/RC/2289\_P2} (Insight SFI Research Centre for Data Analytics), co-funded by the European Regional Development Fund in collaboration with the SFI Insight Centre for Data Analytics at Dublin City University. 

\bibliographystyle{IEEEtran}
\bibliography{references}

\end{document}